# No-Reference Quality Assessment of Contrast-Distorted Images using Contrast Enhancement

Jia Yan, Jie Li, Xin Fu

**Abstract**:
No-reference image quality assessment (NR-IQA) aims to measure the image quality without reference image. However, contrast distortion has been overlooked in the current research of NR-IQA. In this paper, we propose a very simple but effective metric for predicting quality of contrast-altered images based on the fact that a high-contrast image is often more similar to its contrast enhanced image. Specifically, we first generate an enhanced image through histogram equalization. We then calculate the similarity of the original image and the enhanced one by using structural-similarity index (SSIM) as the first feature. Further, we calculate the histogram based entropy and cross entropy between the original image and the enhanced one respectively, to gain a sum of 4 features. Finally, we learn a regression module to fuse the aforementioned 5 features for inferring the quality score. Experiments on four publicly available databases validate the superiority and efficiency of the proposed technique.

## 1. Introduction

As the users' requirements for high-quality images are increasingly rising, no-reference image quality assessment (NR-IQA) has recently become an active research topic, whose objective is to estimate the human perception of image quality without any access to the original reference image [1, 2]. It can be deployed in every location of an image communication system whenever the distorted image is available for evaluation.

The majority of current NR-IQA methods are proposed based on two steps, namely, feature extraction and SVR-based regression module. Natural scene statistics (NSS) based features constitute the majority of existing NR-IQA methods. A plethora of NSS models in different spatial or transform domains have been incorporated into state-of-the-art methods [3-12]. In [12], a bag of feature maps over several transform domains and three color spaces are used, which can be seemed as a combination of the previous methods. Further, a growing body of researchers turn to resorting to advanced neural networks and learning systems, e.g., deep belief net [13, 14], shallow Convolutional Neural Networks (CNN) [15, 16] and deep CNNs [17].

However, traditional NR-IQA studies are mainly devoted to gauging commonly encountered artifacts, for example, Gaussian blur, noise, and compression. In comparison, little investigation is dedicated to the issue of NR-IQA for evaluation of the contrast-changed images. It is worth noting that contrast distortion is often introduced during the image acquisition due to limitation of the acquisition device or poor lighting condition. Moreover, while the aforementioned NR-IQA methods work quite well on commonly seen distortions, they always perform ineffectively on contrast distortion [9], as will also be shown later in this paper. As a matter of fact, IQA of contrast-distorted images is an important research topic, which can be used as a guider or evaluation criteria in contrast enhancement technologies [18, 19].

Currently, there are several studies only focusing on contrast distorted images. Gu et al. [19, 20] first systematically studied this issue; and designed the CID2013 and CCID2014 databases dedicated to image contrast change, and meanwhile proposed reduced reference IQA (RR-IQA) techniques based on phase congruency and information statistics of the image histogram. Liu et al. [21] also proposed a RR-IQA metric RCIQM with the integration of free energy theory and the histogram comparison of the contrast-altered image with that of the reference image. But these methods rely on partial access to the reference image, and therefore, NR algorithms are eagerly required. Fang et al. [22] proposed a NR quality metric based on the NSS regulation including the moment and entropy features, and it achieves better results than the general-purpose NSS based methods. Gu et al. [23] proposed a training-free blind quality method NIQMC based on the concept of information maximization by computing entropy of particular areas and also comparing the image histogram with the uniformly distributed histogram. Gu et al. [24] lately extracted 17 features based on contrast, sharpness, brightness and more, and then yields an IQA metric BIQME using a regression module, which is learned with gathered big-data training samples.

In this paper, we develop a simple yet effective NR-IQA framework for contrast-distorted images by finding that a high contrast image always looks more similar to its enhanced version. Based on this concern, the design philosophy of our blind metric lies in two factors, namely, image similarity, histogram based entropy. Given the input image and the one created through classical histogram equalization (HE), we first concentrate on their appearance. In fact, it is exactly a full-reference IQA problem, and we use classical structural similarity index (SSIM) [25] as the metric. On the other hand, on finding that the histograms of high-contrast and low-contrast images are different, we turn to histograms of the above two images and compute the entropy of original histogram and the enhanced one respectively to reflect the distribution status. Further, we compute the cross entropy to include the interaction between two histograms. Afterward, with the total 5 features, we develop the contrast enhancement based contrast-changed image quality measure (CEIQ) via Support vector regression. Actually, the entropy has already been used for IQA in RCIQM [21] and NIQMC [23]. It is worthwhile to note that CEIQ is different from the previous work. First, the metric RCIQM is not a NR method and needs the histogram of the reference image to compare, while as a NR method, NIQMC compares the input image histogram with the uniformly distributed one. Second, we take advantage of SSIM, which is a benchmark FR-IQA method, in our NR-IQA metric CEIQ because we convert the NR-IQA problem to be a FR-IQA problem by creating a ‘reference image’ via contrast enhancement. At last, we provide a general framework and any similarity measurement function or contrast enhancement method can be employed in it.

The remainder of this paper is organized as follows. In Section II, we present in details the CEIQ method and demonstrate the high relevance of similarities with image contrast. In Section III, thorough experiments verify the superiority and efficiency of our metric. Section IV concludes this paper.

## 2. Similarity measurement with equalization

The feature extraction flowchart of the proposed CEIQ is illustrated in Fig. 1. It consists of the following main steps: decolorization, histogram equalization, image similarity, histogram based entropy and cross entropy computation, which are described as follows.

Given an input color image $\mathrm{I}$, on the first step we convert it to a grayscale image $\mathrm{I}_g$ for subsequent processing. In particular, the concern here is the choice of the decolorization method. Recently, many methods of color-to-grayscale conversion have been proposed [27-30]. However, most of these methods aim to maximally preserve or even enhance the original color contrast [31]. In our CEIQ framework, our goal is to evaluate the real contrast quality, not to produce the visually-pleasing grayscale result. With this, the original contrast of $\mathrm{I}$ should be maintained after decolorization. We produce the grayscale image $\mathrm{I}_g$ by YIQ color model:

$$\mathrm{I}_g = 0.2989 \times \mathrm{I}_{red} + 0.5870 \times \mathrm{I}_{green} + 0.1140 \times \mathrm{I}_{blue} \tag{1}$$

where $\mathrm{I}_{red}$, $\mathrm{I}_{green}$ and $\mathrm{I}_{blue}$ denote the RGB channels of $\mathrm{I}$. The YIQ model uses three components: luminance Y and two values I and Q characterizing the chromaticity. This model is used in the American NTSC television norm, and we take the Y channel as the grayscale representation. It should be noted here that the simple method defined in Eq. 1 could not achieve the best grayscale image, but the result image can preserve the original contrast quality with respect to the color image, which is exactly our goal in CEIQ framework. Fig. 2 shows four color images with different contrast quality level selected from CCID2014 database [19], and the quality increases from left to right. Our results using Eq. 1 are shown in the second column. The results via [29] and [30] which are two famous methods for producing visually-pleasing grayscale images, are shown in the third and fourth rows. It bears out the fact that after decolorization, our results reflect very well contrast orders presented in the original images and the gap between the first and fourth images is large, while the quality levels are less discriminable in Fig. 2(c) and Fig. 2(d).

The basis of the proposed method is on the assumption that a high-contrast image is often more similar to its result after contrast enhancement. Our concern here is the choice of contrast enhancement method. Generally, the mainstream enhancement methods can be classified into two categories: histogram-based and Retinex-based [32, 33]. In this paper, the simplicity and automation (means it performs well on almost all types of images without parameter tuning) are two factors to be considered. One key issue is then which method could be used in designing CEIQ? In spite of excessive contrast enhancement, histogram equalization (HE) has proved to be a simple and effective image contrast enhancement technique by flattening and stretching the dynamic range of the image's histogram [32]. With this, we take advantage of the classical HE to achieve the contrast enhancement. To illustrate the results of HE, some original color images $\mathrm{I}_c$ with different contrast, their grayscale images $\mathrm{I}_g$ and HE results $\mathrm{I}_e$ are shown in Fig. 3 (a-c). As stated above, in our CEIQ framework, any improvement of conventional HE based method can also be used. Unlike the contrast enhancement methods which only focus on the appearance of the result image, we care both the original and result images, more specifically, the similarity of them.

Therefore, given two images $I_g$ and $I_e$, we continue to extract our contrast-aware features in two aspects.

*2.1 Image similarity*

Intuitively, the similarity functions used in many existing FR-IQA methods can all be used to compute inter-image similarity. In this paper, we adopt the SSIM, which is calculated by:

$$SSIM(I_g, I_e) = L(I_g, I_e)^{\alpha} * C(I_g, I_e)^{\beta} * S(I_g, I_e)^{\lambda} \tag{2}$$

where constants $\alpha$, $\beta$ and $\lambda$ mediate the relative importance of the three components [25]. *L*, *C* and *S* measure the similarities of luminance, contrast and structure respectively between $I_g$ and $I_e$. In this work, we follow [25] for the configuration of the parameters. Therefore, we use the similarity of two images defined in Eq. 2 as the first feature $S_{ge}$ in our CEIQ framework.

In our work, we also consider the feature-similarity (FSIM) index [34], which is another excellent FR-IQA method based on SSIM. Its performance will be examined in the section of experimental results.

For the five images shown in Fig. 3, the corresponding mean opinion scores (MOS) increase from left to right, which means that the right one has the highest subjective score given by human. Using Eq. 2, we can easily obtain the corresponding $S_{\mathrm{ge}}$ scores: 0.1457, 0.5364, 0.5562, 0.6404 and 0.7735 respectively, which are illustrated in Fig. 4. It can be realized that $S_{\mathrm{ge}}$ is monotonically related to the MOS score. Further, we exhibit the scatter plots of MOS/DMOS versus $S_{\mathrm{ge}}$ on the whole CCID2014 (including 655 images) and CSIQ (including 116 contrast changed images) database [34] in Fig. 5. Clearly, the score of image similarity shows higher linearity and monotonicity on CCID2014 database, but the performance becomes worse on CSIQ database. So, only one feature is not enough to handle all the contrast distortion models and levels, we should explore more contrast-aware features.

*2.2 Histogram based entropy and cross entropy*

The second consideration of our proposed metric comes from the histogram. The major reason is that histogram is the indicator of the contrast. We plot the corresponding histograms in Fig. 3(d) and Fig. 3(e). More precisely, entropy is a chief concept in statistics [26], and it represents the amount of information for a random signal by quantifying its average unpredictability. It is reasonable to assume that not only the high-contrast image, but also its contrast enhanced version are of the histogram toward uniform distribution, which stands for the large entropy. Given the histogram $h$ of pixel values, the entropy $E$ is defined by

$$E = -\sum_{i=0}^{b} h(i) \log h(i) \tag{3}$$

where $b$ indicates the value of bins in histogram and $\log(\cdot)$ has base of two. Therefore, given two histograms $h_g$ and $h_e$ corresponding to the input gray image and the equalized image, we can easily obtain their entropy $E_g$ and $E_e$ by Eq. 3 as our second and third features.

Further, in our work, we take the relationship between $h_g$ and $h_e$ into consideration by the use of the cross entropy, which is computed by:

$$E_{ge} = -\sum_{i=0}^{b} h_g(i) \log h_e(i) \tag{4}$$

It should be noted here that the combination of Eq. 3 and Eq. 4 is exactly the classical forma of K-L divergence to estimate the distance between two distributions. However, we take the individual part as one feature in our work to describe the characteristic of one histogram and the interaction between two histograms separately. In the same way, another cross entropy is computed by:

$$E_{eg} = -\sum_{i=0}^{b} h_e(i) \log h_g(i) \tag{5}$$

Thus, we obtain our third and forth features.

Finding a good tradeoff between efficacy and efficiency, we divide the histogram into 128 bins when computing, and we find that the value of bins (such as 64, 128 or 256) has little effect on the performance of CEIQ. In addition, we only take the bins with non-zero values to compute the above four features.

In Fig. 6, two scatter plots of DMOS versus $E_e$ and $E_{eg}$ on the CSIQ (including 116 contrast changed images) database are given. From them, we can see that these features perform well than the image similarity on CSIQ database. So, we use the above described five features to represent the contrast quality of the input image.

*2.3 The CEIQ metric*

For any image, we utilize the image similarity $s_{ge}$, histogram based entropy $E_g$ and $E_e$, cross entropy $E_{ge}$ and $E_{eg}$ to form the feature vector $f$. We can construct a training set of $k$ images with their feature vectors and subjective scores $\{(f_1, Q_1), (f_2, Q_2), \cdots, (f_k, Q_k)\}$. A regression function could be learned to find the mapping function.

The framework is generic enough to allow for the use of any regressor. In our implementation, a SVM regressor (SVR) is adopted. SVR has previously been applied to IQA problems [4, 7, 10, 11, 12, 20, 22, 24]. In this work, we use the LIBSVM package [36] with the linear kernel to implement the SVR. Once the regression model is learned, we use it to estimate the quality of any input image.

## 3. Experimental Results and Discussions

### *3.1 Databases and methods for comparison*

The source code of our CEIQ has been released at *https://github.com/mtobeiyf/CEIQ*. The proposed method is examined on four contrast distortion related image databases: CID2013 [20], CCID2014 [19], CSIQ [34] and TID2013 [37]. CID2013 and CCID2014 are specifically designed for contrast distortion, including 400 and 655 contrast-changed images generated from 15 natural images respectively. CSIQ is built for various distortion types, among which there are 116 global contrast changed images. TID2013 is a large scale image quality database which contains 3000 images generated from 25 reference with 24 distortion types for each reference. A total number of 240 images which are directly related to contrast distortions (#16 mean shift and #17 contrast change) are utilized here.

We compare the proposed CEIQ with eight state-of-the-art IQA methods: 1) classical FR-IQA methods, including SSIM [25], FSIM [34] and MAD [38]; 2) representative general-purpose NR-IQA methods, including BRISQUE [4] and IL-NIQE [9]; 3) contrast distortion specific NR-IQA methods, including Fang [22], NIQMC [23] and BIQME [24]. It should be stressed here that BRISQUE, FANG, BIQME and our CEIQ are the methods need training, and we put them into a group. Specially, we also consider the score of $s_{\mathrm{ge}}$ as one metric to evaluate its performance, and it should be stressed here that this is a training-free metric. The performance of the involved methods is evaluated by three popular criteria computed between predicted scores and subjective scores: Spearman rank order correlation coefficient (SROCC) and Kendall rank order correlation coefficient (KROCC) for prediction monotonicity, Pearson linear correlation coefficient (PLCC) for prediction accuracy. PLCC is computed after the monotonic logistic mapping between objective and subjective scores [39]. A value close to 1 for SROCC, KROCC and PLCC means superior correlation with subjective opinions.

### *3.2 Overall performance comparison*

For the training-free methods, we use all the images in each database to obtain the results. Taking the TID2013 for example, 250 images are all used to compute the coefficients. For the methods require training, we split each database into train and test sets with 80% for training and 20% for testing. The split is conducted according to reference images to guarantee the independency of the image content in training set and test set. This splitting is repeated for 1,000 times and the median results are used to evaluate the final performance. This treatment method is also widely used in training based IQA research, like the experiments in [4, 22, 24].

Table I lists the SROCC, PLCC and KROCC values of examined methods on four databases to show their performance. For better comparing, we divide the NR methods into two groups: training-free and training based. From the results, five conclusions can be derived. First, among training-free methods, even FR ones, our $s_{\mathrm{ge}}$ achieves promising results compared to other methods on CID2013, CCID2014 and TID2013 databases. This demonstrate that the image similarity of the input image and the equalized one is really a powerful feature. Second, our CEIQ metric is obviously superior to all the training based models on CCID 2014, CSIQ and TID2013 databases, especially on CSIQ and TID2013 where the CEIQ outperforms others by a significant margin. As to the CID2013 database, the CEIQ is comparable with the best one BIQME (0.8934 vs. 0.9015 with SROCC). Third, apparently, the contrast distortion specific methods outperform the general-purpose methods (IL-NIQE and BRISQUE) by a large extent because of the special selection of contrast-aware features. Fourth, our CEIQ performs more accurately and consistently than our $S_{\text{ge}}$ on all the databases. There are two reasons: 1) entropy based features are added for training step means the model can 'study' the difference between distortion levels; 2) the smaller images are included in the evaluation (20% vs. 100%). Fifth, we surprisingly find that our $S_{\text{ge}}$ , which is a training-free metric, even outperforms some training based methods on some databases. For example, the SROCC of $S_{\text{ge}}$ on CID2013 is 0.8431, while the result of FANG is 0.7852. The results of $S_{\text{ge}}$ on the four databases can also be found at *https://github.com/mtobeiyf/CEIQ* .

Since our CEIQ is a training based metric, which requires distorted images and their MOS to learn the model, we report the performance of it under different training set sizes. 10%-80% of the images in each database are used for training and the left ones are used for test according to the same method described previously. For each ratio, we compute the median SROCC values among 1000 times on each database, and the results are shown in Fig. 7. This figure witnesses slightly fluctuations from 10% to 80%. In fact, by comparing the results in Table I and Fig. 7, even with only 10% of images for training, the SROCC results of our metric are also quite encouraging compared with other metrics.

In the above experiments, the training and test samples are drawn from the same database. It is expected that the BIQA model learned from one database should be applicable to images in other databases. Therefore, we test the generalization capability of our metric by training it on one database, then testing it on other databases. The SROCC score is used for evaluation, and the results are presented in Fig. 8. Note that when trained on CID2013 or CCID2014 and then applied to other database, our metric deliver excellent performance and outperform other metrics which are trained and test in the same database in Table I. However, it performs poorly when trained on CSIQ or TID2013 database and test on CID2013 and CCID2014 database. The major reason is that the number of images and the contrast distortion levels of CID2013 and CCID 2014 are more than them of the CSIQ and TID2013. For example, there are 45 distorted images generated from one reference image on average in CCID2014 database, while only 4 distorted images corresponding to one reference image are available in CSIQ database.

### *3.3 Runtime measure*

To test the efficiency which is also important in real-world applications, the runtime of involved NR metrics are listed in the Table II. Results are performed on a notebook with Intel Core i7-6700HQ CPU@2.6G Hz and 8GB RAM. The software platform is Matlab R2017a. The computational cost is evaluated with $768 \times 512$ RGB image from CCID2014 database. Clearly, our proposed CEIQ owns highest efficiency, followed by BRISQUE and FANG. Compared to the features extracted in BIQME which performs also well in the previous experiments, the major cost of our metric is just the SSIM computation. Hence, the proposed method is quite efficient, outperforming BIQME by a large margin.

### *3.4 Similarity measurement strategies*

As described above, our proposed metric illustrated in Fig. 1 provides a general framework for IQA of contrast-distorted images, and any available method can be employed in it, such as the similarity measurement strategies. Besides the SSIM which we used, in this section, we consider another classical measurement for inter-image similarity: FSIM [34]. The SROCC results of these methods on CCID2014 database are tabulated in Table III. From the results, it is possible to see that although the FSIM performs better than SSIM in classical FR-IQA problems, its performance is worse than SSIM in our CEIQ. In FSIM, phase congruency (PC) is used as the primary feature [34]. However, PC is contrast invariant, meaning that the images with different contrast distortion level have almost the same scores for PC, which is the main reason of the worse performance of FSIM in our framework.

## 4 Conclusion

In this paper, we develop a no-reference quality metric CEIQ to assess the quality of contrast changed images using histogram equalization from two aspects: image similarity and histogram based entropy and cross entropy. Experimental results show that the proposed methods are well correlated with subjective evaluations even only few images are used for training, suggesting that the proposed metrics are excellent at handling contrast-distorted images with very low computational cost. Further, CEIQ provides a general framework, and other contrast enhancement methods and image similarity measure metrics can be employed in this framework.

**Acknowledgments**

This work is supported by the National Nature Science Foundation of China (No. 61701351).

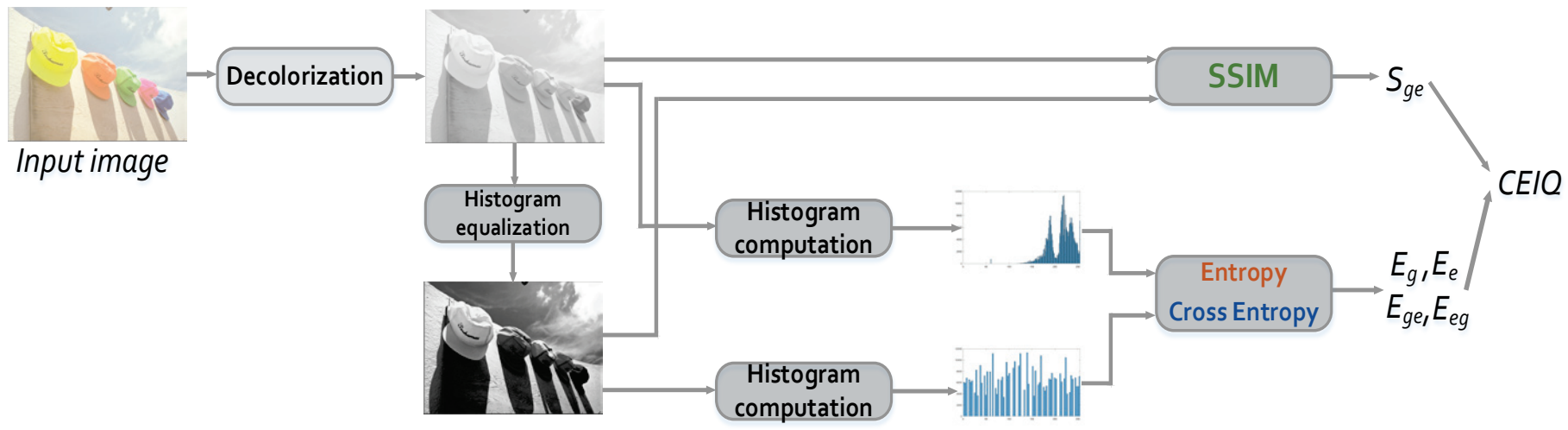

Fig. 1. The feature extraction flowchart of our proposed CEIQ algorithm

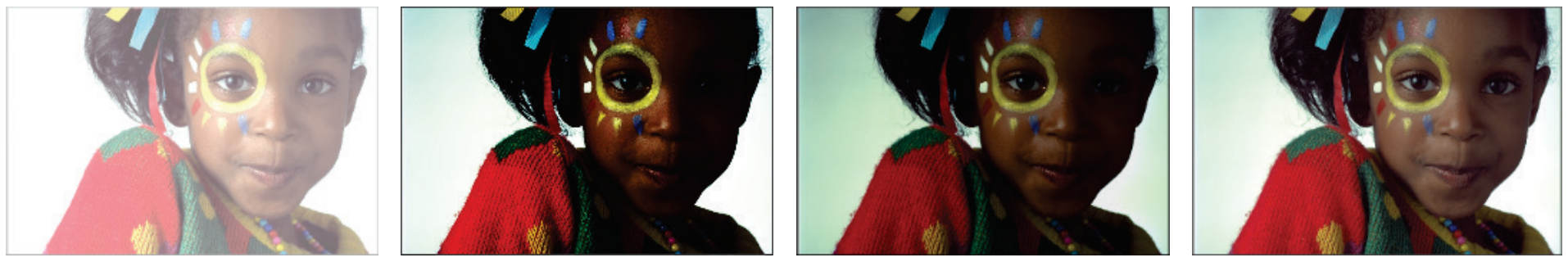

MOS=2.6044 MOS=2.7124 MOS=3.1133 MOS=3.4956

(a) Color images

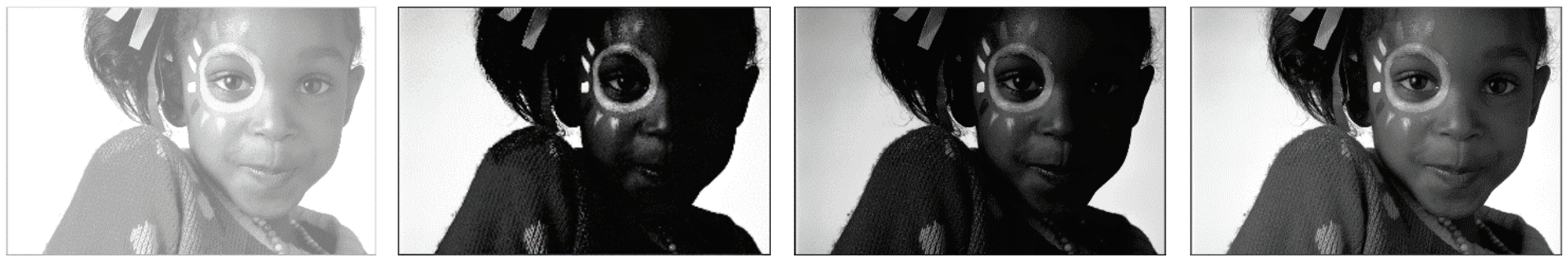

(b) Grayscale images generated by Eq. 1

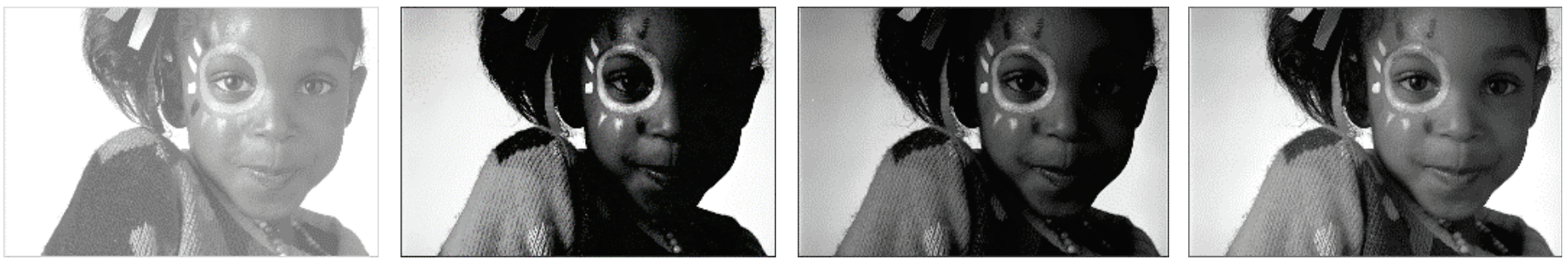

(c) Grayscale images generated by 'Decolorize' [10]

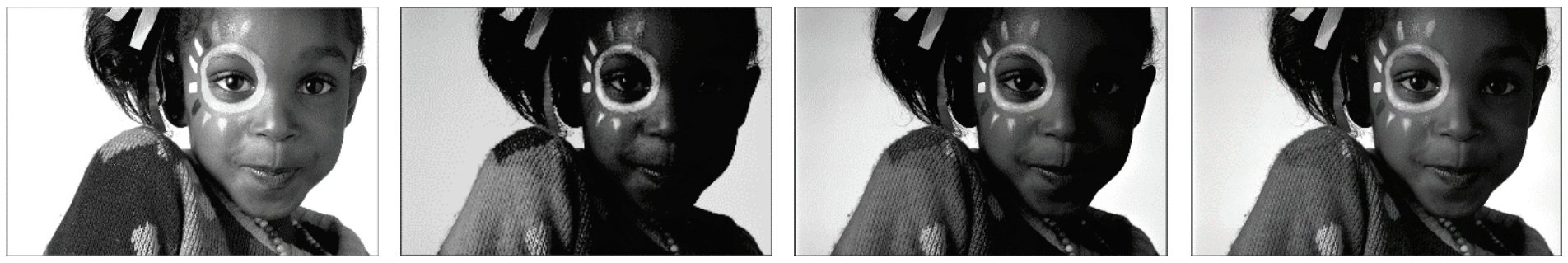

(d) Grayscale images generated by 'CPD' [11]

Fig. 2. Comparison with other decolorization methods

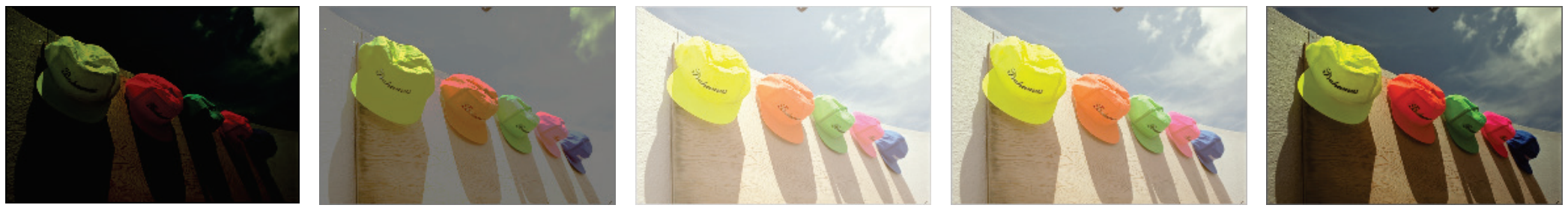

(a) Color images $\mathbf{I}$

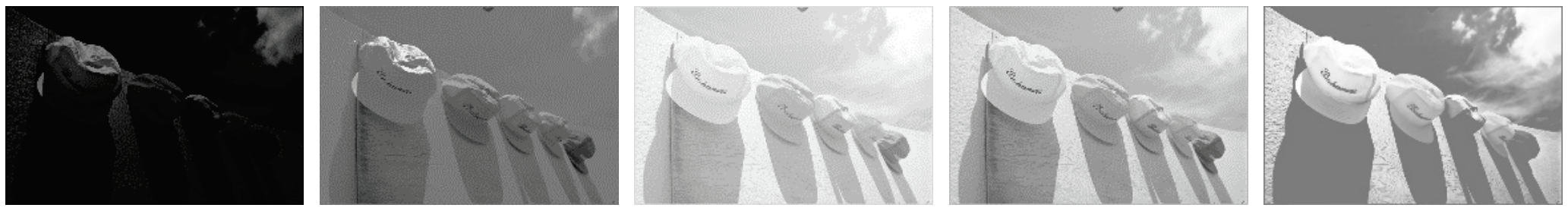

(b) Grayscale images $\mathbf{I}_g$

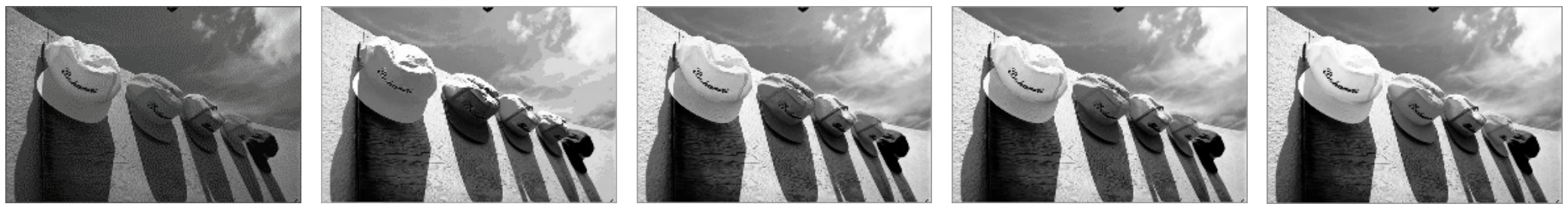

(c) Contrast enhanced images $\mathbf{I}_e$

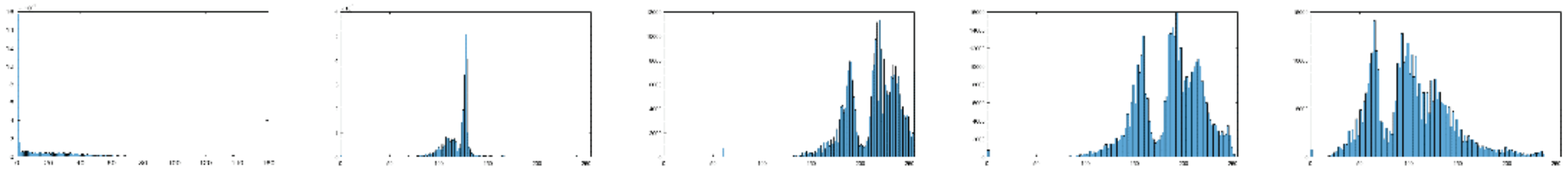

(d) Histogram $\mathbf{h}_g$ of $\mathbf{I}_g$

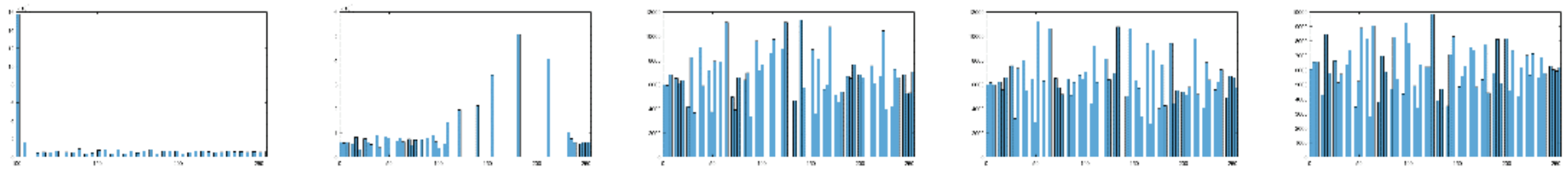

(e) Histogram $\mathbf{h}_e$ of $\mathbf{I}_e$

Fig. 3. The histograms of grayscale images and the contrast enhanced ones using HE.

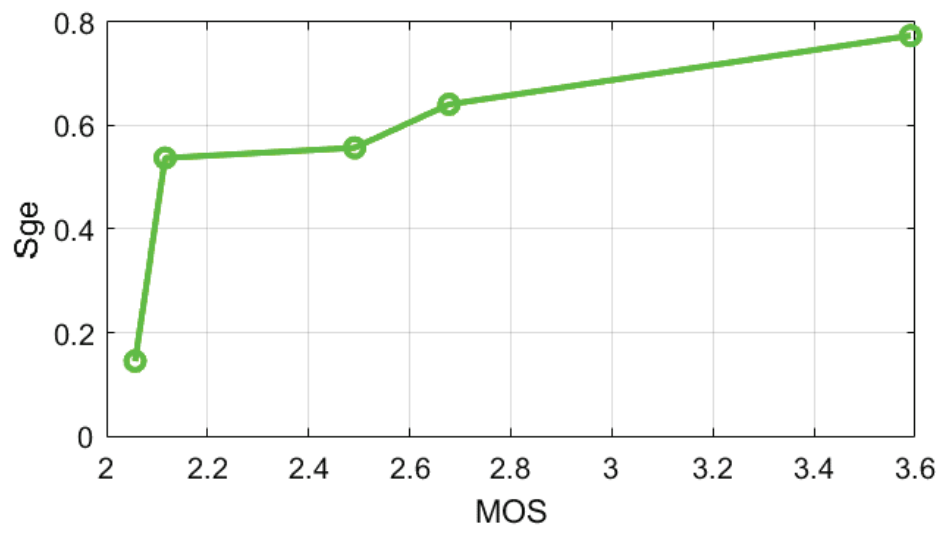

Fig. 4. Comparison between $S_{ge}$ and subjective MOS of five images from CCID2014

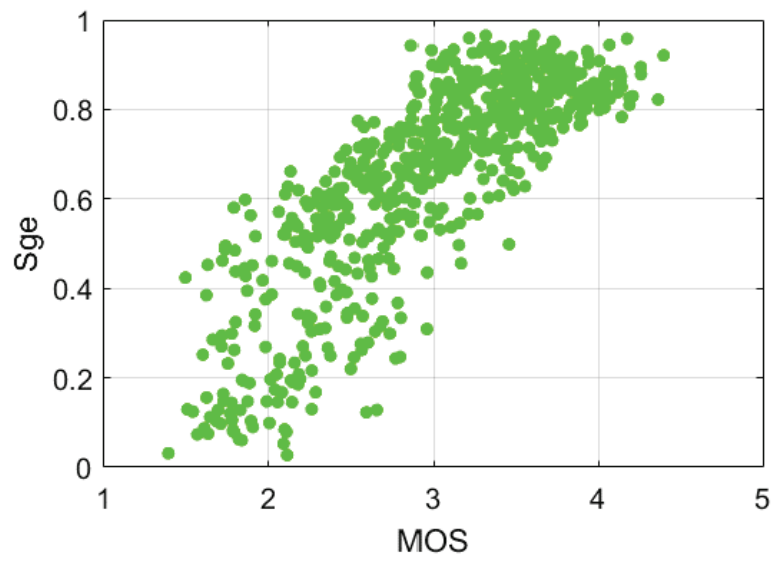

(a)

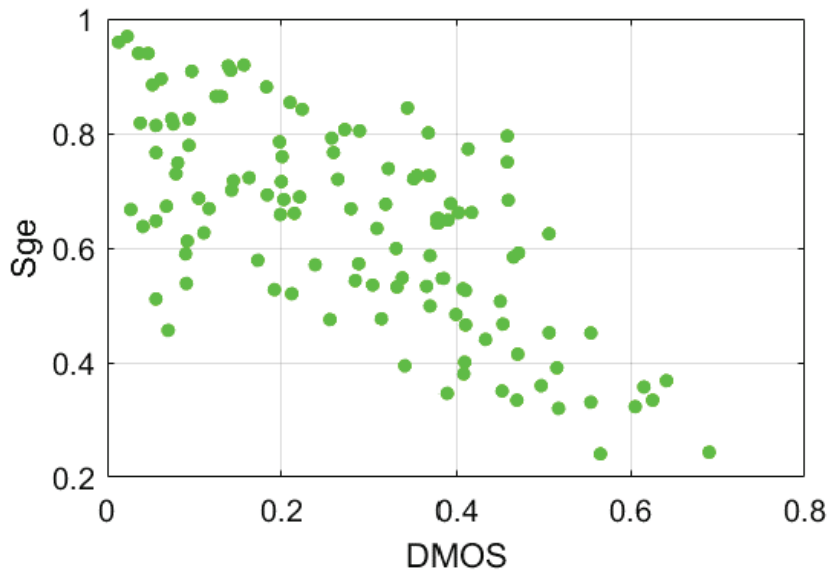

(b)

Fig. 5. Scatter plots of MOS/DMOS versus $S_{ge}$. (a) CCID2014 database (b) CSIQ database

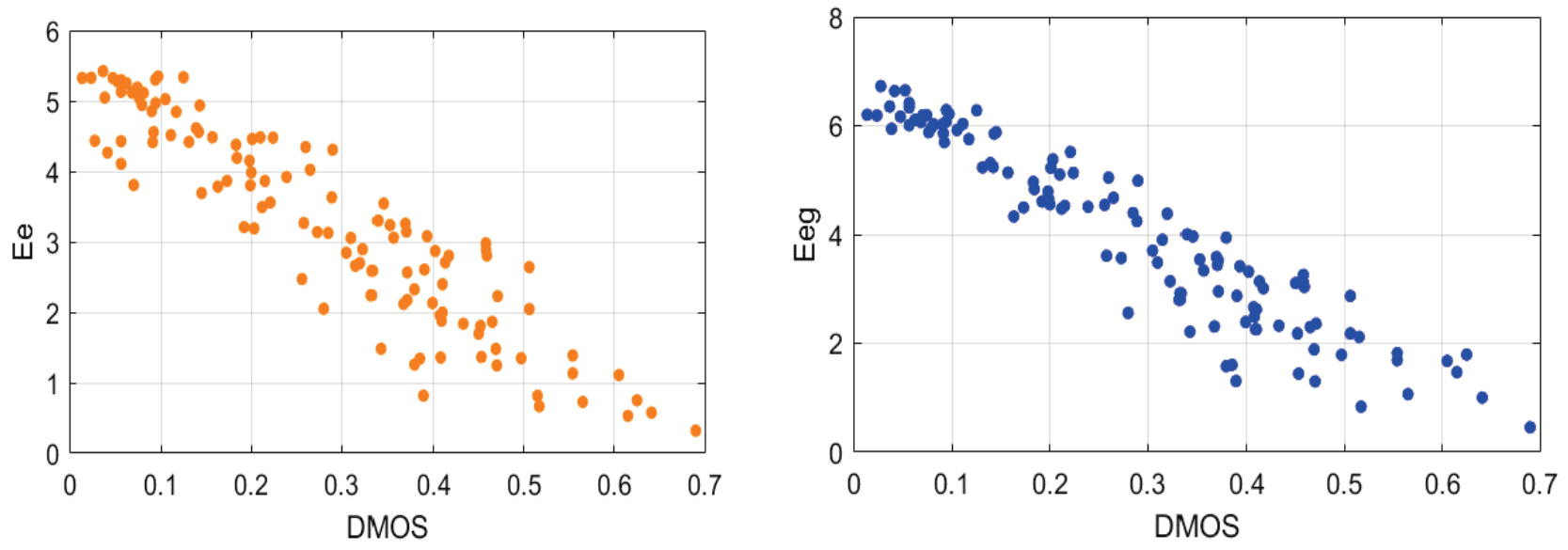

(a) (b)

Fig. 6. Scatter plots of DMOS versus $E_e$ (a) and $E_{eg}$ (b) on CSIQ database

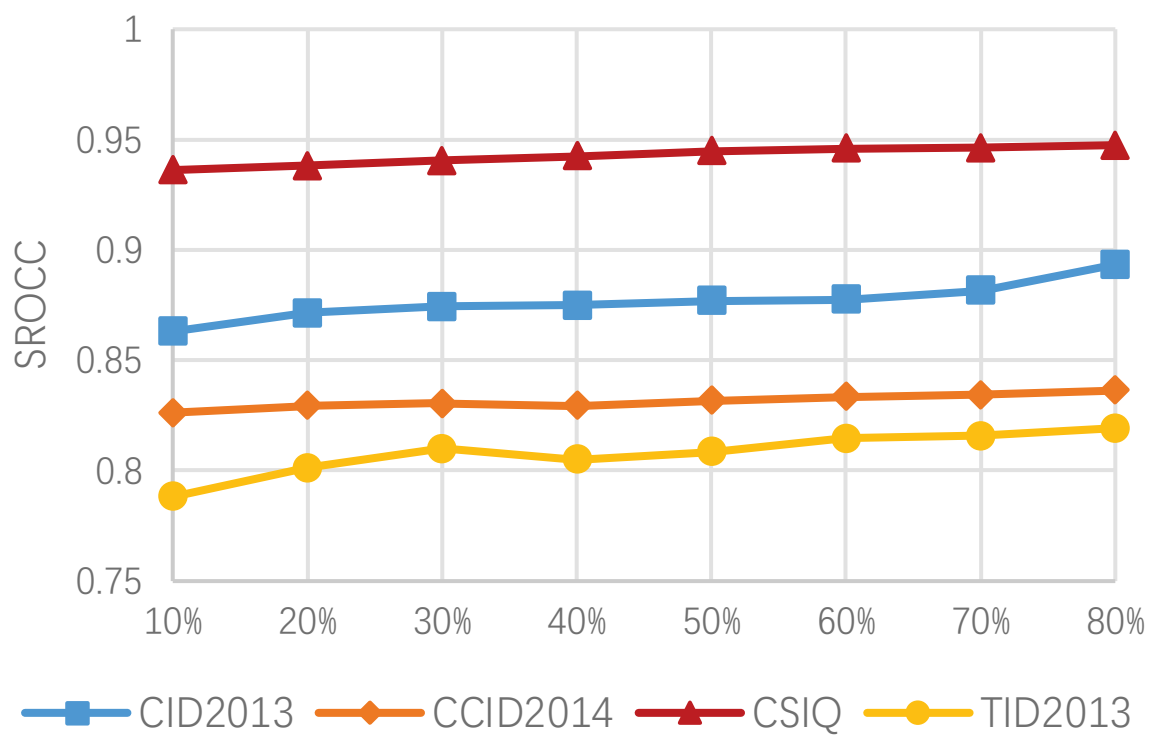

Fig. 7. SROCC with different split ratios on four image databases

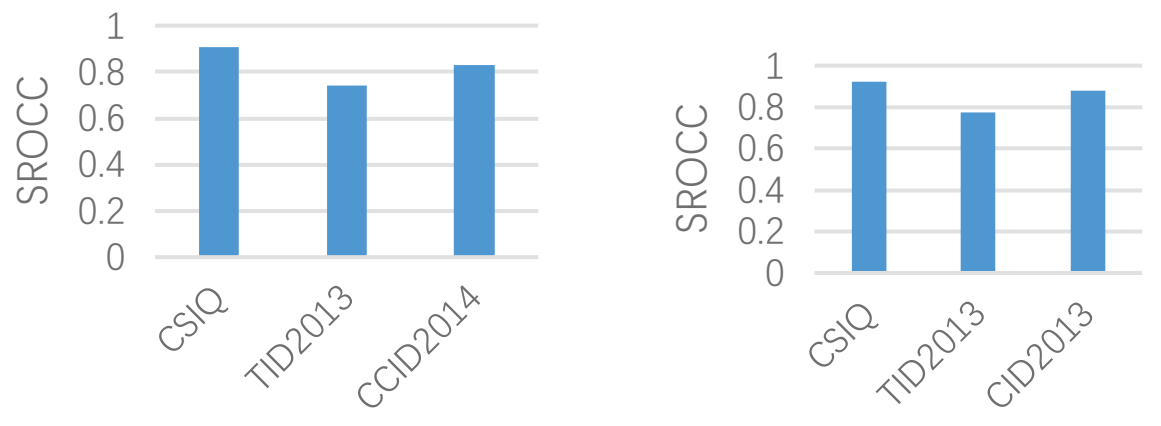

(a) Trained on CID2013 (b) Trained on CCID2014

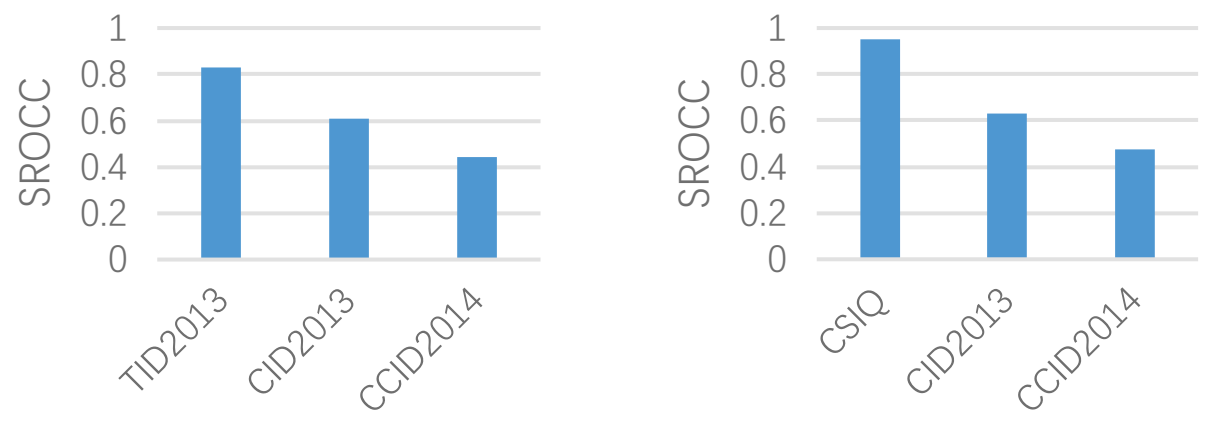

(c) Trained on CSIQ (d) Trained on TID2013

Fig. 8 Cross-Database performance of our proposed metric

TABLE I

Performance comparison of 10 IQA methods. The top one for each group is highlighted.

| Quality Metrics | Type | Training | CSIQ (116 images) | | | TID2013 (240 images) | | |
|---|---|---|---|---|---|---|---|---|
| | | | SROCC | PLCC | KROCC | SROCC | PLCC | KROCC |
| SSIM | FR | No | 0.7397 | 0.7450 | 0.5323 | **0.4905** | 0.5658 | 0.3432 |
| FSIM | FR | No | **0.9420** | **0.9378** | **0.7883** | 0.4413 | **0.6819** | **0.3588** |
| MAD | FR | No | 0.9207 | 0.9321 | 0.7460 | 0.3300 | 0.4077 | 0.2558 |
| IL-NIQE | NR | No | 0.5005 | 0.5468 | 0.3510 | 0.1517 | 0.2275 | 0.1030 |
| NIQMC | NR | No | **0.8533** | **0.8747** | **0.6689** | 0.6458 | **0.7225** | **0.4687** |
| $S_{ge}$(Pro.) | NR | No | 0.6154 | 0.7112 | 0.4483 | **0.6491** | 0.6029 | 0.4442 |
| BRISQUE | NR | Yes | 0.2539 | 0.3488 | 0.1706 | 0.0551 | 0.1429 | 0.0359 |
| FANG | NR | Yes | 0.7232 | 0.6998 | 0.5178 | 0.3478 | 0.4113 | 0.2291 |
| BIQME | NR | Yes | 0.8202 | 0.8276 | 0.6501 | 0.6510 | 0.7479 | 0.4783 |
| CEIQ(Pro.) | NR | Yes | **0.9475** | **0.9532** | **0.8182** | **0.8193** | **0.8718** | **0.6302** |

| Quality Metrics | Type | Training | CID2013 (400 images) | | | CCID2014 (655 images) | | |
|---|---|---|---|---|---|---|---|---|
| | | | SROCC | PLCC | KROCC | SROCC | PLCC | KROCC |
| SSIM | FR | No | 0.8132 | 0.8072 | 0.5513 | **0.8136** | **0.8256** | **0.6063** |
| FSIM | FR | No | **0.8486** | **0.8574** | **0.6663** | 0.7658 | 0.8201 | 0.5707 |
| MAD | FR | No | 0.8079 | 0.8151 | 0.5912 | 0.7430 | 0.7928 | 0.5458 |
| IL-NIQE | NR | No | 0.5273 | 0.5682 | 0.3708 | 0.5121 | 0.5764 | 0.3590 |
| NIQMC | NR | No | **0.8668** | **0.8691** | **0.6690** | 0.8113 | **0.8438** | **0.6052** |
| $S_{ge}$(Pro.) | NR | No | 0.8431 | 0.8524 | 0.6412 | **0.8120** | 0.8422 | **0.6052** |
| BRISQUE | NR | Yes | 0.2552 | 0.3351 | 0.1745 | 0.2123 | 0.3575 | 0.1445 |
| FANG | NR | Yes | 0.7852 | 0.7801 | 0.6035 | 0.7786 | 0.8012 | 0.5711 |
| BIQME | NR | Yes | **0.9015** | **0.8989** | **0.7323** | 0.8321 | 0.8511 | **0.6565** |
| CEIQ(Pro.) | NR | Yes | 0.8934 | 0.8960 | 0.7085 | **0.8363** | **0.8675** | 0.6362 |

TABLE II

Runtime comparison of evaluated NR metrics

| Metrics | Number of features | Time (second/image) |
|---|---|---|
| IL-NIQE | 468 | 2.331 |
| NIQMC | 2 | 2.213 |
| BRISQUE | 36 | 0.211 |
| FANG | 5 | 0.5831 |
| BIQME | 17 | 0.762 |
| CEIQ | 5 | 0.056 |

TABLE III

SROCC results of different combination forms in CEIQ

| Metrics | SROCC |
| --- | --- |
| CEIQ (SSIM) | 0.8363 |
| CEIQ (FSIM) | 0.8121 |